\documentclass{article}
\usepackage{ijcai23}

\usepackage{times}
\usepackage{soul}
\usepackage{url}
\usepackage[hidelinks]{hyperref}
\usepackage[utf8]{inputenc}
\usepackage[small]{caption}
\usepackage{graphicx}
\usepackage{amsmath}
\usepackage{amsthm}
\usepackage{booktabs}
\usepackage{algorithm}
\usepackage{algorithmic}
\usepackage[switch]{lineno}

\usepackage{algorithm}
\usepackage{algorithmic}
\usepackage[utf8]{inputenc} 
\usepackage[T1]{fontenc}    
\usepackage{url}            
\usepackage{booktabs}       
\usepackage{amsfonts}       
\usepackage{nicefrac}       
\usepackage{microtype}      
\usepackage{xcolor}         
\usepackage{graphicx}
\usepackage{float}
\usepackage{subfig}
\usepackage{overpic}
\usepackage{amsmath}
\usepackage{enumitem}

\usepackage{newfloat}
\usepackage{listings}

\DeclareMathOperator*{\argmin}{arg\,min}

\title{Efficient Multi-View Inverse Rendering Using a Hybrid Differentiable\\ Rendering Method}

\author{
Xiangyang Zhu$^{1,2}$
\and
Yiling Pan$^{1,2}$\and
Bailin Deng$^{3}$\And
Bin Wang$^{1,2}$\footnote{Corresponding Author}
\affiliations
$^1$School of Software, Tsinghua University, China\\
$^2$Beijing National Research Center for Information Science and Technology (BNRist), China\\
$^3$School of Computer Science and Informatics, Cardiff University, UK\\
\emails
\{zhuxy20, pyl16\}@mails.tsinghua.edu.cn,
DengB3@cardiff.ac.uk,
wangbins@tsinghua.edu.cn
}

\begin{document}

\maketitle

\begin{abstract}
Recovering the shape and appearance of real-world objects from natural 2D images is a long-standing and challenging inverse rendering problem. In this paper, we introduce a novel hybrid differentiable rendering method to efficiently reconstruct the 3D geometry and reflectance of a scene from multi-view images captured by conventional hand-held cameras. Our method follows an analysis-by-synthesis approach and consists of two phases. In the initialization phase, we use traditional SfM and MVS methods to reconstruct a virtual scene roughly matching the real scene. Then in the optimization phase, we adopt a hybrid approach to refine the geometry and reflectance, where the geometry is first optimized using an approximate differentiable rendering method, and the reflectance is optimized afterward using a physically-based differentiable rendering method. Our hybrid approach combines the efficiency of approximate methods with the high-quality results of physically-based methods.  Extensive experiments on synthetic and real data demonstrate that our method can produce reconstructions with similar or higher quality than state-of-the-art methods while being more efficient.
\end{abstract}

\begin{figure*}[t]
  \includegraphics[width=\textwidth]{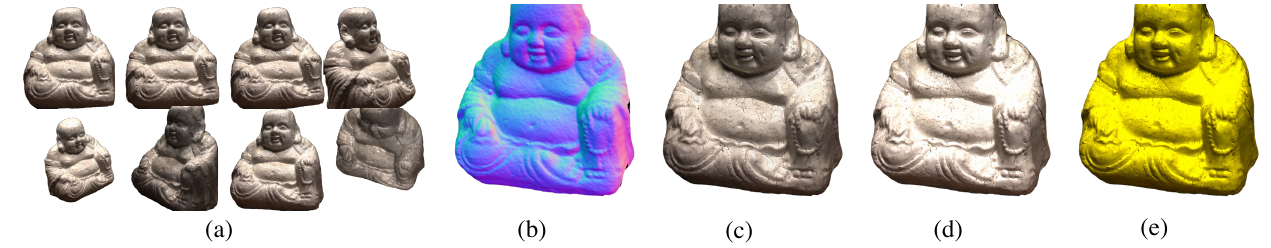}
  \caption{(a) Our method takes as input a set of images obtained by conventional hand-held cameras from several viewpoints and gets a rough initial model (b) by traditional methods. Then, we perform a novel analysis-by-synthesis optimization to refine the model’s shape and reflectance separately, yielding a high-quality 3D model. In (c) and (d), we show a re-rendering of the result under a novel viewpoint and environmental lighting. In addition, we can edit the material (e).}
  \label{firstpage}
\end{figure*}

\section{Introduction}

Inverse rendering, which recovers 3D geometry, reflection properties and illumination from 2D images, is a long-standing and challenging problem in computer graphics and vision~\cite{patow2003survey}. Benefiting from the advances in deep neural networks, many deep learning-based methods  learn to obtain the materials and plane normals of near-flat objects in a data-driven way~\cite{li2018materials,aittala2016reflectance,gao2019deep}. However, these methods struggle to deal with complex geometry, resulting in limited application in practice. Some other deep learning-based methods use 3D geometric representations such as signed distance fields~\cite{yariv2020multiview,zhang2021physg}, tetrahedral meshes~\cite{munkberg2021extracting}, and occupancy functions~\cite{mildenhall2020nerf,zhang2021nerfactor,boss2021nerd,srinivasan2021nerv,zhang2021ners,zhang2022iron}, to handle more complex geometries. However, such geometry representations may require post-processing in order to be used in traditional graphics pipelines~\cite{lorensen1987marching,remelli2020meshsdf}, which may cause material information loss and affect the rendering image quality.

The emergence of differentiable rendering techniques~\cite{kato2020differentiable} means that image loss can be back-propagated along the rendering pipeline to solve inverse rendering problems, which promotes the development of multi-view inverse rendering tasks using triangular meshes as the geometry representation. Approximate differentiable rendering methods \cite{loper2014opendr,kato2018neural,liu2019soft,chen2019learning}  utilize simplified rendering processes and can solve the inverse rendering problem efficiently,  but the simple material models they use usually lead to poor visual effects. On the contrary, physically-based differentiable rendering methods 
\cite{li2018differentiable,li2021shape} can reconstruct high-quality physically-based rendering (PBR) materials in the path tracing manner, but their realistic results come at a high computational cost. Although some methods have been proposed to accelerate the reconstruction process~\cite{luan2021unified}, they often impose strong assumptions on the lighting conditions to narrow down the parameter search space, which hinders their wide application. 

In this paper, we propose a hybrid differentiable rendering method to reconstruct triangular meshes and PBR materials from multi-view real-world images captured by conventional hand-held cameras (\emph{e.g.} mobile phones) with uncontrolled lighting conditions. 
Our method consists of two phases.
In the initialization phase, we use traditional methods to reconstruct a rough triangular mesh for the scene. Then in the optimization phase, we take a hybrid approach to optimize the scene geometry and PBR materials, where we first optimize the geometry using an approximate method, followed by the PBR materials using a physically-based method.
Our novel formulation benefits from both the efficiency of approximate methods and the high quality of physically-based methods. 
Extensive experiments show that our method achieves a significantly faster training and rendering speed than state-of-the-art methods, while achieving results of comparable or better quality.

In summary, our contributions include:
\begin{itemize}[leftmargin=*]
\item We propose a novel hybrid differentiable rendering optimization method based on triangular meshes, which utilizes an approximate differentiable rendering method to optimize the geometry, and a physically-based differentiable rendering method to get the PBR materials.
\item 
The proposed pipeline is user-friendly and can work end-to-end. Furthermore, the optimized scene parameters can be easily deployed on mainstream commercial graphics engines without the need for conversion, making it applicable to a wide range of applications such as virtual reality and augmented reality.
\item 
We conduct extensive experiments on both synthetic and real-world data, which demonstrate that our method outperforms or achieves comparable results to the state-of-the-art methods while being significantly more efficient.

\end{itemize}

\begin{figure*}[t]
	\includegraphics[width=\textwidth]{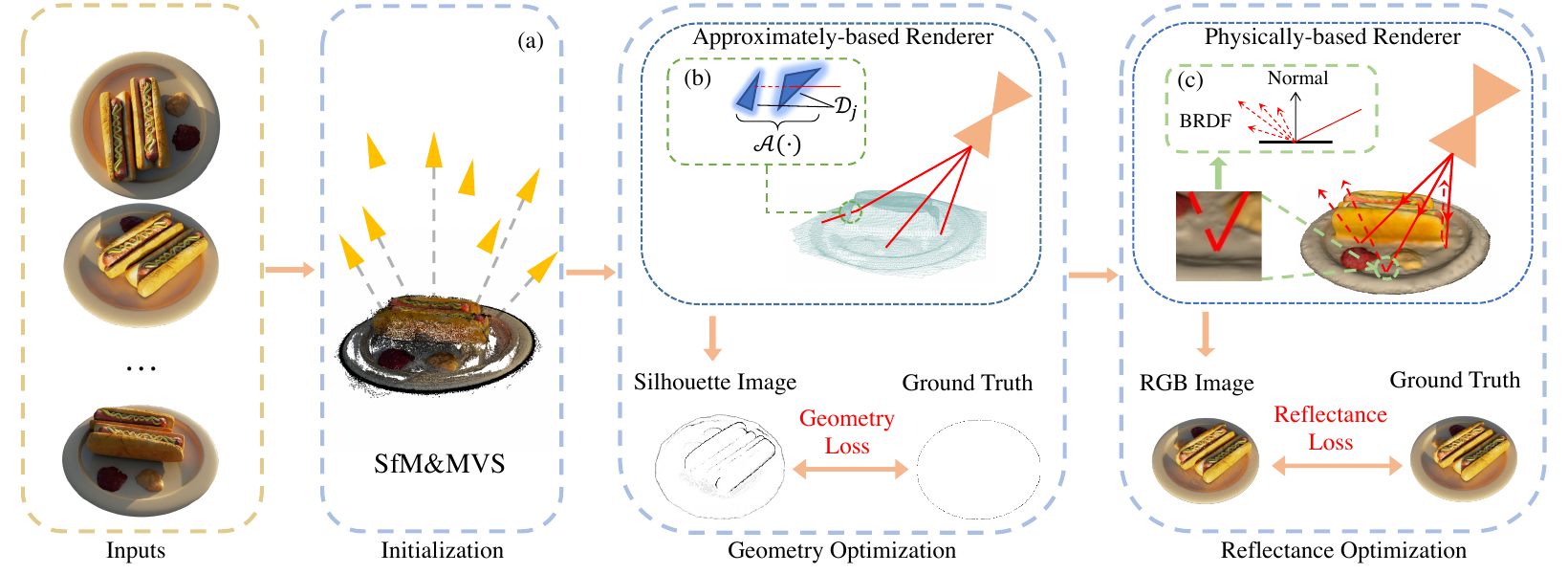}
	\caption{Overview of our inverse rendering pipeline. Our method takes as input a set of RGB images of some object obtained by conventional hand-held cameras from several viewpoints with uncontrolled lighting conditions. Then, we utilize SfM and MVS method to reconstruct a virtual scene roughly matching the real scene in the initialization phase. In the optimization phase, we first optimize the geometry using the approximately-based differentiable rendering method, then optimize reflection using the physically-based differentiable rendering method. Both methods work iteratively by pushing the loss between rendering images and the ground truth in the camera pose.}
	\label{pipeline}
\end{figure*}

\section{Related Work}


\paragraph{Shape Reconstruction.}
Reconstructing object geometry is  a long-standing problem in computer vision and graphics. 
In traditional methods, the Structure-from-Motion (SfM) method \cite{schonberger2016structure} is first applied to generate sparse feature points and find their correspondence to further generate a sparse point cloud and a rough mesh. Then, the Multi-View-Stereo (MVS) method \cite{schonberger2016pixelwise} is leveraged to generate dense pixel-level matching. Finally, a dense mesh with vertex color is generated by the Poisson reconstruction method \cite{kazhdan2013screened}. A few recent learning-based methods assume that the target object mesh is homeomorphic with the sphere. The method in~\cite{wang2018pixel2mesh} uses an image feature network (2D CNN) to extract perceptual features from the input image, and a cascaded mesh deformation network (GCN) to progressively deform an ellipsoid mesh into the desired 3D model.  These methods can only reconstruct rough geometry, which is insufficient for downstream tasks such as VR and AR.

\paragraph{Reflectance Reconstruction.}
Reflectance model, \emph{i.e.}, spatially-varying bidirectional reflectance distribution function (SVBRDF), describes how light interacts with surfaces in the scene. 
Traditional SVBRDF reconstruction methods \cite{matusik2003data,lensch2003image,holroyd2010coaxial,dong2010manifold,chen2014reflectance,dong2015predicting,kang2018efficient} rely on dense input images measured using auxiliary equipment, \emph{e.g.}, gantry. Some other works focus on \cite{zhou2016sparse,kim2017lightweight,park2018surface} exploiting the structure of SVBRDF parameter spaces to reduce the requirement for the number of input images. Additionally, some data-driven works \cite{li2018materials,aittala2016reflectance,gao2019deep} have been introduced recently to produce plausible SVBRDF estimations for near-flat objects using a small number of input images. Despite their ease of use, these methods struggle to handle more complex objects.

\paragraph{Differentiable Rendering.}
Differentiable methdos are reviewed in detail in~\cite{kato2020differentiable}. Here we focus on methods closely related to our work.
Traditional rendering pipelines, \emph{e.g.}, rasterization and ray-tracing, are not differentiable due to some discrete parts, which means that they cannot work with the gradient descent method just like neural networks. The emergence of differentiable rendering has changed all. Some approximately-based methods have been proposed recently.  \cite{loper2014opendr,kato2018neural,kato2019learning} compute approximated gradients to optimize scene parameters. Besides, \cite{liu2019soft,chen2019learning} replace the z-buffer-based triangle selection of a vanilla rasterization process with a probabilistic manner. Unfortunately, the result is not so good due to inaccuracies introduced by these methods. On the contrary, Monte Carlo edge sampling based methods \cite{li2018differentiable,zhang2019differential} provide unbiased gradient estimates capable of producing plausible results. But these methods are resource-consuming because of their path tracing module, hindering their generalization.



\paragraph{Differentiable Rendering based Multi-view Inverse Rendering.}
The emergence of differentiable rendering boosts the development of the inverse rendering techniques. Several prior works leverage differentiable rendering methods to solve the inverse problem.
Wu et al.~\shortcite{wu2020unsupervised} propose a completely unsupervised method for face reconstruction from just one face image using the approximately-based differentiable method, which does not rely on existing 3D morphable face models. Luan et al.~\shortcite{luan2021unified} leverage the Monte Carlo edge sampling based methods to reconstruct fine geometry and plausible material. However, they assume that the camera and light are collocated. In addition, a rough geometry scanned by professional equipment is also needed. Although these settings could narrow down the solution space, they also reduce the generalization of the method. The method proposed by Li et al.~\shortcite{li2021shape}  reduces the requirements for input. It takes as input multi-view wild scene images, and reconstructs the initial geometry through MVS, then use the general Monte Carlo path tracing differentiable renderer to optimize the material, illumination, and geometry. Although it can achieve good results, it is resource-consuming and time-consuming.

\section{Our Method}

\subsection{Overview}

Our inverse rendering pipeline, shown in Figure~\ref{pipeline}, takes as input a set of RGB images captured by conventional hand-held cameras from multiple viewpoints. Using traditional methods (SfM and MVS), our method first reconstructs a virtual scene (in the form of a triangle mesh with vertex color) that roughly matches the real scene in the initialization phase. Afterward, in the optimization phase, our method first uses an approximate differentiable rendering method to optimize the geometry, then uses a physically-based differentiable rendering method to optimize the reflectance. Both methods iteratively improve the image loss between rendered images and the ground truth for the same viewpoint. Finally, the optimized scene parameters can be used for a variety of applications such as novel view synthesis and relighting. Conceptually, our pipeline can be formulated as an analysis-by-synthesis problem
\begin{equation*}
    \Theta^{*}= \argmin_{\Theta} \mathcal{L}(\mathcal{I}(\Theta), \Theta ; \tilde{\mathcal{I}}),
\end{equation*}
where $\Theta$ represents the scene parameters that need to be estimated, including geometry and reflection parameters;
$\mathcal{L}$ is a loss function to be minimized;
$\tilde{\mathcal{I}}$ is the ground truth image, and $\mathcal{I}(\Theta)$ is the image rendered by our hybrid differentiable rendering method using the same camera parameters as $\tilde{\mathcal{I}}$.
In the following, we present the details of each phase.

\subsection{Initialization Phase}


Our reconstruction pipeline is designed to work in complex, unstructured scenes where the input images have no additional information, such as depth. To tackle these challenges, we leverage traditional methods to obtain an initial geometry. Specifically, as shown in Figure~\ref{pipeline}(a), we use the SfM method from~\cite{schonberger2016structure} to generate sparse feature points in each image. These points are then used to reconstruct a sparse point cloud and a rough mesh by matching the feature points across images. Next, we use the MVS reconstruction method from~\cite{schonberger2016pixelwise} to generate dense pixel-level matching. Finally, a dense mesh is then obtained from Poisson surface reconstruction~\cite{kazhdan2006poisson}, with vertex colors derived from the input images. 


In terms of lighting, we use an environment map as our lighting source. The resolution of the environment map is $512\times 128$. The environment map is a learnable parameter that is inferred during the material optimization phase. In the beginning, we assume that the light is white and the three channels of our environment map are set to $(0.5,0.5,0.5)$.

\subsection{Optimization Phase}

In the optimization phase, starting from the initial geometry obtained in the previous step, we adopt a hybrid differentiable rendering method to further optimize the geometry and reflectance. Specifically, we utilize an approximate differentiable rendering method to optimize the geometry, and a physically-based differentiable rendering method to optimize the reflectance. 
Our hybrid approach is motivated by the following observations in experiments:
\begin{itemize}[leftmargin=*]
\item Although the geometry reconstructed by SfM and MVS in the initialization phase provides a good approximation of the scene, there can be some defects in the boundary regions.
\item Approximate differentiable rendering methods dedicated to calculating the gradient of geometry can be very efficient, but the visual effect of their results may be of lower quality due to their simplification of the material model.
\item 
Physically-based methods can accurately reconstruct complex materials, but their high computational costs and resource demands may impede their efficiency.
\end{itemize}
Specifically, starting from the triangle mesh obtained in the initialization phase, we further optimize its vertex positions $\theta_\mathrm{p}$ which define the geometry, as well as two 2D texture maps that contain the diffuse albedo $\theta_{\mathrm{d}}$ and specular albedo $\theta_{\mathrm{s}}$ respectively and define the SVBRDF. In this way, our scene parameters can be represented as $\mathbf{\Theta}=\left(\theta_{\mathrm{p}}, \theta_{\mathrm{d}}, \theta_{\mathrm{s}}\right)$.
We first optimize the vertex positions using an approximate differentiable rendering method. Afterward, we optimize the diffuse albedo and specular albedo using a physically-based differentiable rendering method. Details of the optimization are presented below.


\paragraph{Geometry Optimization.}
To optimize the geometry, we adopt a differentiable model similar to Soft Rasterizer~\cite{liu2019soft} to  compute a silhouette image for each input image using the same camera parameters (see the supplementary materials for details of the computation), to indicate the occupancy from the same view directions as the input images. Then we minimize the following loss function to derive the vertex positions $\theta_\mathrm{p}$:
\begin{equation}
    \mathcal{L}_{\text{geo}} = \lambda_{\text {sil}}\mathcal{L}_{\text {sil }}
    + \lambda_{\text {lap}}\mathcal{L}_{\text {lap }}+\lambda_{\text {normal}}\mathcal{L}_{\text {normal }}+\lambda_{\text {edge}}\mathcal{L}_{\text {edge }},
    \label{Lgeo}
\end{equation}
where $\mathcal{L}_{\text {sil }}$ is a silhouette loss that indicates the consistency between the computed silhouette images  $\mathcal{I}_{\text{sil}}(\theta_\mathrm{p})$ and the ground-truth ones $\tilde{\mathcal{I}}_{\text{sil}}$ derived from the input~\cite{liu2019soft}:
\[
\mathcal{L}_{\text {sil}}=1-{\left\|\tilde{\mathcal{I}}_{\text{sil}} \otimes \mathcal{I}_{\text{sil}}\right\|_{1}}/{\left\|\tilde{\mathcal{I}}_{\text{sil}} \oplus \mathcal{I}_{\text{sil}}-\tilde{\mathcal{I}}_{\text{sil}}\otimes \mathcal{I}_{\text{sil}}\right\|_{1}},
\]
with $\otimes$ and $\oplus$ being the element-wise product and sum operators, respectively.
The other terms in $\mathcal{L}_{\text{geo}}$ are regularizers. Among them, $\mathcal{L}_{\text{lap}}=\|\textbf{\emph{L}}\textbf{\emph{V}}\|^{2}$ is a mesh Laplacian loss $\mathcal{L}_{\text {lap }}$ of a mesh with $n$ vertices, $\textbf{\emph{V}}$ is the $n \times 3$ coordinates matrix, and $\textbf{\emph{L}} \in \mathbb{R}^{n \times n}$ is the Laplacian matrix of the mesh (See \cite{nealen2006laplacian} for details).
$\mathcal{L}_{\text {normal }}=\sum_{i, j}\left[1-\left(\textbf{\emph{n}}_{i} \cdot \textbf{\emph{n}}_{j}\right)\right]^{2}$ is a normal consistency loss to make the normals of adjacent faces to vary slowly, where the sum is over all triangle pairs $(i, j)$ sharing a common edge, and $\textbf{\emph{n}}_i$ and $\textbf{\emph{n}}_j$ are the face normals of the two specific triangles.
$\mathcal{L}_{\text {edge }}=\sqrt{\sum_{i} e_{i}^{2}}$ is an edge length loss to avoid long edges that can cause ill-shaped triangles, where $e_{i}$ denotes the length of the $i$-th edge.



\paragraph{Reflectance Optimization.}
Our reflectance optimization is based on the rendering equation proposed in~\cite{kajiya1986rendering}. 
For a surface point $x$ with surface normal $n$, let $L_i(\omega_i; x)$ be the incident light intensity at location $x$ along the direction $\omega_i$, and SVBRDF $f_r(\omega_o, \omega_i; x)$ be the reflectance coefficient of the material at location $x$ for incident light direction $\omega_i$ and viewing direction $\omega_o$. Then the observed light intensity $L_o(\omega_o;x)$ is an integral over the upper hemisphere $\Omega$:
\begin{equation*}
L_o(\omega_o;x)=\int_{\Omega}L_i(\omega_i)f_r(\omega_o,\omega_i;x)(\omega_i \cdot n)d_{\omega_i}.
\end{equation*}
We leverage a modified Cook-Torrance (CT) model \cite{cook1982reflectance} based on~\cite{Zeltner2021MonteCarlo} as our reflectance model, which contains a rough surface with diffuse and specular reflection without refraction. Mathematically, our reflectance model can be described as the following,
\begin{equation*}
    f_{r}\left(\omega_o, \omega_i; x\right)=\theta_{d}(x)+\theta_{s}\left(\omega_o, \omega_i; x\right),
\end{equation*}
where $\theta_d$ and $\theta_s$ are the diffuse and specular reflectance:
\begin{align*}
    \theta_{d}(x) & ={\rho_{d}(x)}/{\pi},\\
    \theta_{s}\left(\omega_o, \omega_i; x\right) &=\rho_{s}(x) \frac{D(h, \alpha) G\left(\omega_o, \omega_i, n(x)\right)}{\left(n(x) \cdot \omega_i\right)\left(n(x) \cdot \omega_o\right){\pi}}.
\end{align*}
 Specifically, $\rho_d$ is diffuse albedos, $\rho_s$ is specular albedos, and $h$ is the halfway vector. $D(h, \alpha)$ is the microfacet distribution function which is GGX \cite{walter2007microfacet} used in our work. $\alpha$ and $n(x)$ denote the surface's roughness and normal, respectively. $G$ is a shadowing-masking function. Like other works, we also ignore the Fresnel effect which cannot be observed in our scene.
 To optimize the reflectance parameters $\theta_\mathrm{r}=\left(\theta_{\mathrm{d}}, \theta_{\mathrm{s}}\right)$, we minimize the following reflectance loss:
\begin{equation}
    \mathcal{L}_{\text{ref}} =\lambda_{\text {rgb}}\mathcal{L}_{\text{rgb}}(\mathcal{I}_{\text{rgb}}(\theta_\mathrm{r});\tilde{\mathcal{I}}_{\text{rgb}})+\lambda_{\text{reg}}\mathcal{R}(\theta_\mathrm{r}),
    \label{Lref}
\end{equation}
where $\mathcal{L}_{\text {rgb }}={\left\|\tilde{\mathcal{I}}_{\text{rgb}} - \mathcal{I}_{\text{rgb}}(\theta_\mathrm{r})\right\|_{1}}$ measures the $\ell_{1}$ norm of the difference between the rendered color image $\mathcal{I}_{\text{rgb}}(\theta_\mathrm{r})$ and the ground truth $\tilde{\mathcal{I}}_{\text{rgb}}$.
$\mathcal{R}(\theta_\mathrm{r})$ is a regularizer for the diffuse albedo $\theta_\mathrm{d}$ and specular albedo $\theta_\mathrm{s}$. Similar to~\cite{schmitt2020joint}, we assume that nearby pixels with similar diffuse albedos have similar specular albedos. So we choose 
$$\mathcal{R}(\theta_\mathrm{r}) = \sum\nolimits_{\textbf{\emph{p}}} \left\| \theta_{\mathrm{s}}[\textbf{\emph{p}}]-\left(\sum\nolimits_{\textbf{\emph{q}}} \theta_{\mathrm{s}}[\textbf{\emph{q}}] k_{\textbf{\emph{p}}, \textbf{\emph{q}}}\right) /\left(\sum\nolimits_{\textbf{\emph{q}}} k_{\textbf{\emph{p}}, \textbf{\emph{q}}}\right) \right\|_{1},$$ 
where $k_{\textbf{\emph{p}}, \textbf{\emph{q}}}=\exp \left(-\frac{\|\textbf{\emph{p}}-\textbf{\emph{q}}\|_{2}^{2}}{2 \sigma_{1}^{2}}- \frac{\left(\theta_{\mathrm{d}}[\textbf{\emph{p}}]-\theta_{\mathrm{d}}[\textbf{\emph{q}}]\right)^{2}}{2 \sigma_{2}^{2}}\right)$ and $\textbf{\emph{p}}, \textbf{\emph{q}}$ are two specific mesh vertices.

\begin{figure*}[t]
	\centering
	\includegraphics[width=0.95\textwidth]{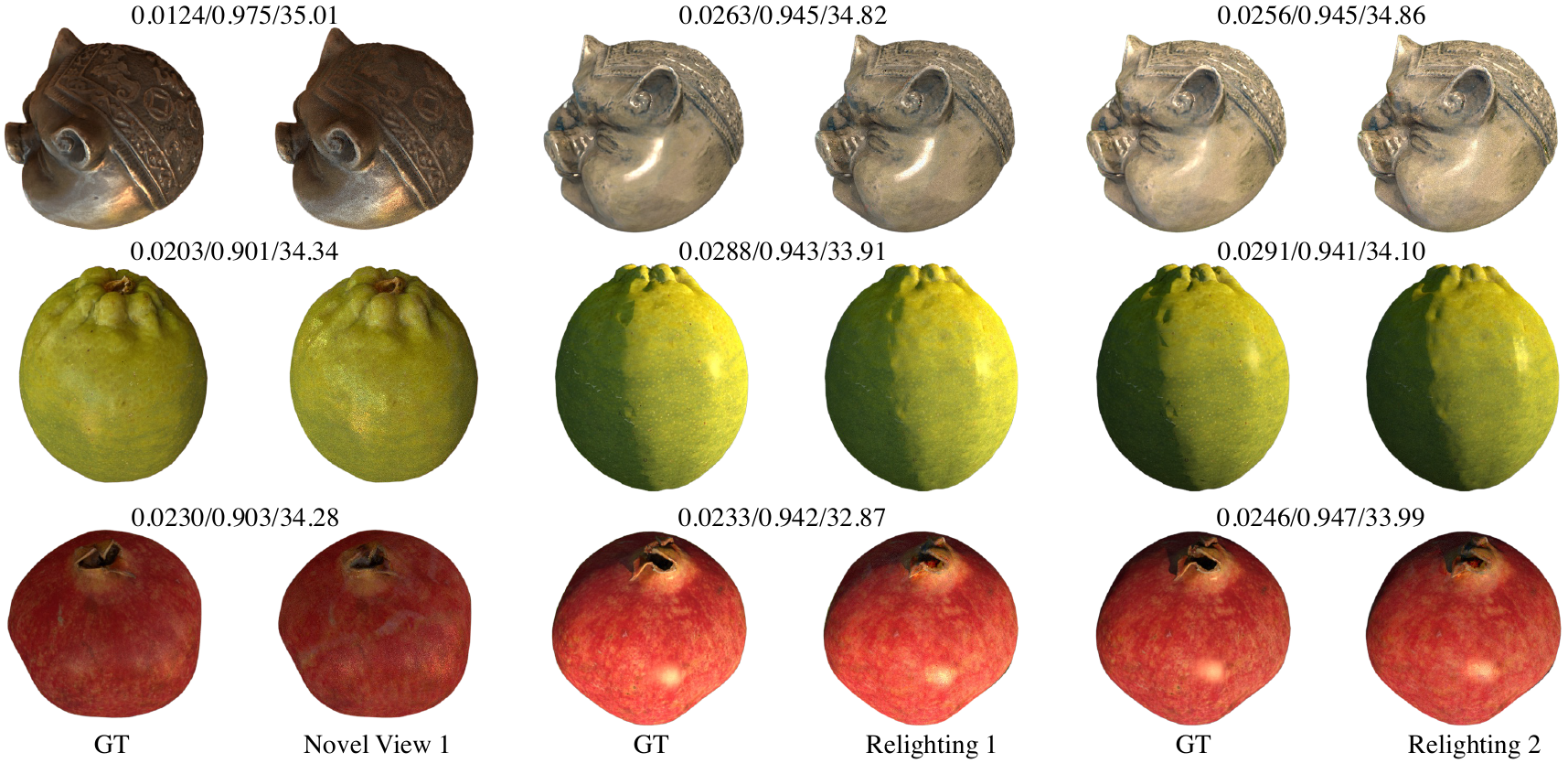}
	\caption{Results of our method on the synthetic data. We compare our predicted novel views and relighting results to the ground truth images. The number above each result indicates LPIPS, SSIM, and PSNR metrics calculated in $512\times512$ size pictures, respectively. \textbf{Please zoom in to see the details, especially the challenging specular highlights on the surface of the objects.}}
	\label{synthesis-graph}
\end{figure*}

\begin{figure*}[t]
	\centering
	\includegraphics[width=\textwidth]{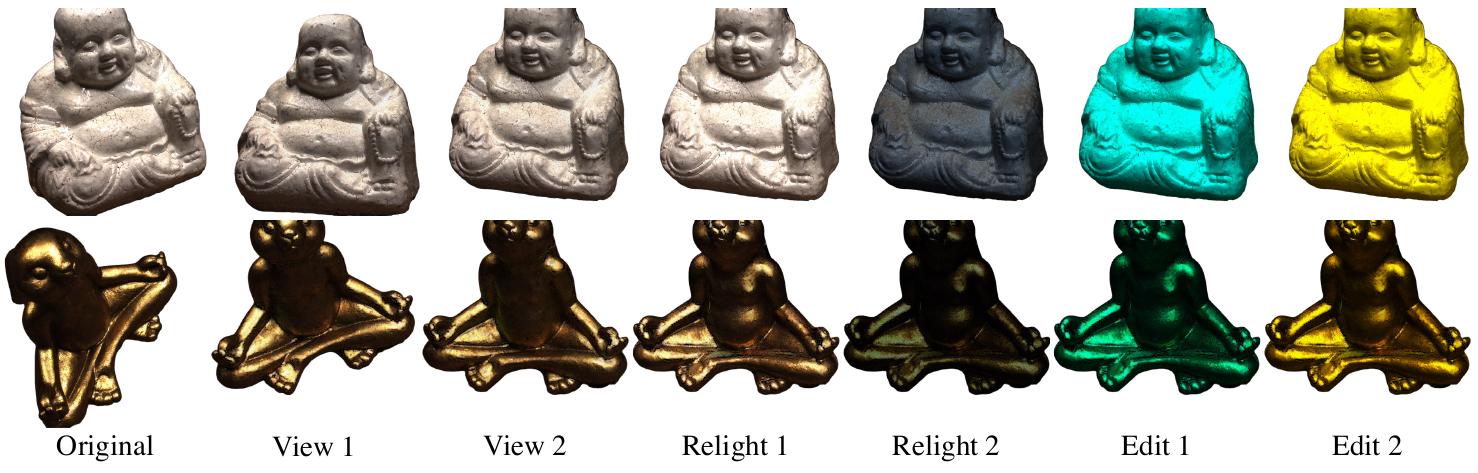}
	\caption{With our pipeline, we can synthesize novel views and edit the materials and lighting of the real-world captures.}
	\label{3}
\end{figure*}

\section{Experiments}

We perform quantitative and qualitative evaluation of our method using extensive experiments on both synthetic and real data, and compare it with state-of-the-art methods.

\paragraph{Implementation Details.}
We implement the geometry reconstruction module  in the initialization phase using COLMAP\footnote{\url{https://github.com/colmap/colmap}} \cite{schonberger2016pixelwise,schonberger2016structure}, a general-purpose Structure-from-Motion and Multi-View Stereo pipeline. In the optimization phase, we implement the approximate differentiable rendering based on Pytorch3d~\cite{ravi2020accelerating}, and the physically-based differentiable rendering based on Mitsuba~2~\cite{nimier2019mitsuba}. 
The rest of our optimization pipeline is implemented with PyTorch using the Adam optimizer.
For geometry optimization, we use the weights $\left(\lambda_{\text{sil }}, \lambda_{\text{lap }}, \lambda_{\text{edge }}, \lambda_{\text {normal }}\right)=(1.0,1.0,1.0,0.01)$ for Eq.~\ref{Lgeo}, and $0.001$ for the learning rate. 
The ground-truth silhouette images required for Eq.~\ref{Lgeo} are obtained either from rendering (for synthetic data) or using a background removal tool\footnote{\url{https://www.remove.bg/}} (for real data).
For reflectance optimization, we use the weights $\left(\lambda_{\text{rgb}}, \lambda_{\text {reg}}\right)=(0.1,1.0)$ for Eq.~\ref{Lref}, and $0.0001$ for the learning rate. 
To avoid excessive memory consumption in the physically-based rendering module, we set the maximum depth of ray tracing bounce to $3$, the downsampling factor of raw images to $4$, and the number of sampling per pixel $spp $ to $4$ in the iterative optimization phase.

We train and test our method on one NVIDIA RTX 3090 GPU. For geometry optimization, our implementation uses $200$ iterations per view and takes about $0.5$ hours per example. For reflectance optimization phase, our implementation uses $400$ iterations per view and takes $1.5$$\sim$$2.5$  hours per example.


\begin{figure*}[t]
	\centering
	\includegraphics[width=\textwidth]{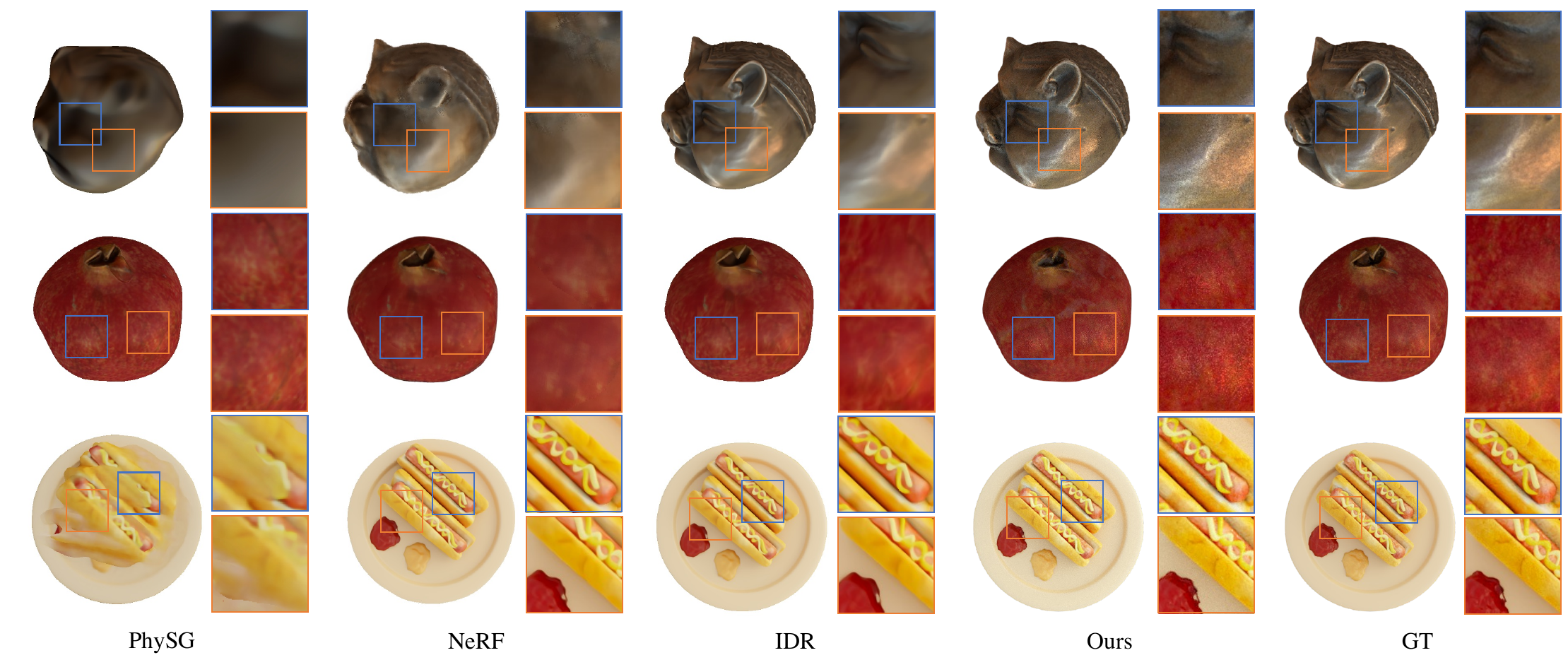}
	\caption{We qualitatively compare our results with PhySG~\protect\cite{zhang2021physg}, NeRF~\protect\cite{mildenhall2020nerf} and IDR~\protect\cite{yariv2020multiview}. }
	\label{comparative}
\end{figure*}

\begin{figure}[h!]
  \centering
  \includegraphics[width=\linewidth]{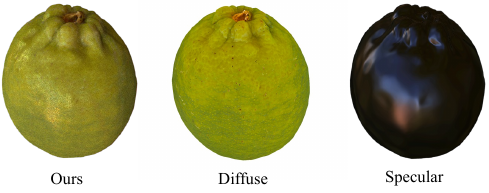}
  \caption{Our diffuse and specular albedo. We have adjusted the tonal range to make the variation of specular albedo more visible.}
  \label{fig:lemon}
\end{figure}

\paragraph{Synthetic Data.}
Our synthetic data is created using meshes and textures from~\cite{zhou2016sparse} and the Internet, covering different materials and geometries. Concretely, for each object, we render 400 images with colored environmental lighting using graphics engines, 300 for training, and the left 100 for novel view synthesis testing, whose viewpoints are evenly distributed inside the upper hemisphere. In addition, we also render the same object with two other environment lighting maps as the ground truth for the following relighting performance testing. 
To quantitatively evaluate our method, we use three image quality metrics\textemdash{}LPIPS~\cite{zhang2018unreasonable}, SSIM and PSNR\textemdash{}to compare the rendering results with the corresponding ground truth images.
Figure~\ref{synthesis-graph} shows examples of results from our method and their ground truth, as well as their evaluation metric values.
Both the quantitative metrics and the qualitative visualization show that our novel views and relighting results match the ground truth closely.
Fig.~\ref{fig:lemon} shows the diffuse and specular albedo for the lemon model.

\begin{table*}[t]
\setlength{\tabcolsep}{1.8pt}
\centering
\resizebox{\textwidth}{!}{
\begin{tabular}{ccccccccccccccccccc}
\toprule 
& &\multicolumn{5}{c}{Synthetic Pig (512$\times$512)} & & \multicolumn{5}{c}{Synthetic Pomegranate (512$\times$512)} & & \multicolumn{5}{c}{Synthetic Hotdog (512$\times$512)}\\
& &\multicolumn{5}{c}{\#train=300, \#test=100} & & \multicolumn{5}{c}{\#train=300, \#test=100} & & \multicolumn{5}{c}{\#train=300, \#test=100}\\
 \cmidrule(lr){3-7}
 \cmidrule(lr){9-13}
 \cmidrule(lr){15-19}
 & &  &  &  & Training & Testing & & & & & Training & Testing & & & & & Training & Testing\\
& & LPIPS$\downarrow$ & SSIM$\uparrow$ & PSNR$\uparrow$ & Time (h)$\downarrow$ & Time (s)$\downarrow$ & & LPIPS$\downarrow$ & SSIM$\uparrow$ & PSNR$\uparrow$ 
& Time (h)$\downarrow$ & Time (s)$\downarrow$ & & LPIPS$\downarrow$ & SSIM$\uparrow$ & PSNR$\uparrow$ & Time (h)$\downarrow$ & Time (s)$\downarrow$\\
\midrule[\heavyrulewidth]
PhySG  & & $0.2200$ & $0.784$ & $17.39$ & $\sim15$ & $\sim4$ & &$0.0425$ & $0.928$ & $23.98$ & $\sim15$ & $\sim4$ & & $0.1473$ & $0.8319$ & $22.30$ & $\sim15$ & $\sim4$\\
NeRF  & & $0.0878$ & $0.859$ & $27.19$ & $\sim15$ & $\sim10$ & & $0.0600$ & $0.918$ & $31.91$ & $\sim15$ & $\sim10$ & &$0.0453$ & $0.921$ & $27.90$ &  $\sim15$ & $\sim10$\\
IDR  & & $0.0183$ & $0.955$ & $29.94$ & $\sim18$ & $\sim5$ & &$\bold{0.0211}$ & $\bold{0.942}$ & $26.85$ & $\sim18$ & $\sim5$ & & $0.0303$ & $0.945$ & $30.97$ & $\sim18$ & $\sim5$\\
Ours & & $\bold{0.0124}$ & $\bold{0.975}$ & $\bold{35.01}$ & $\bold{\sim3}$ & $\bold{\sim1}$ & & $0.0230$ & $0.903$ & $\bold{34.28}$ & $\bold{\sim 2.5}$ & $\bold{\sim1}$ & &$\bold{0.0088}$ & $\bold{0.951}$ & $\bold{33.89}$ & $\bold{\sim3}$ & $\bold{\sim1}$\\
\bottomrule
\end{tabular}
}
\caption{
Comparison between PhySG~\protect\cite{zhang2021physg}, NeRF~\protect\cite{mildenhall2020nerf}, IDR~\protect\cite{yariv2020multiview} and our method.
\#train and \#test denote the size of training set and test set, respectively. Training time is in hours and rendering time is in seconds.
}
\label{comparative-table}
\end{table*}


\begin{figure*}[t]
	\centering
	\includegraphics[width=\textwidth]{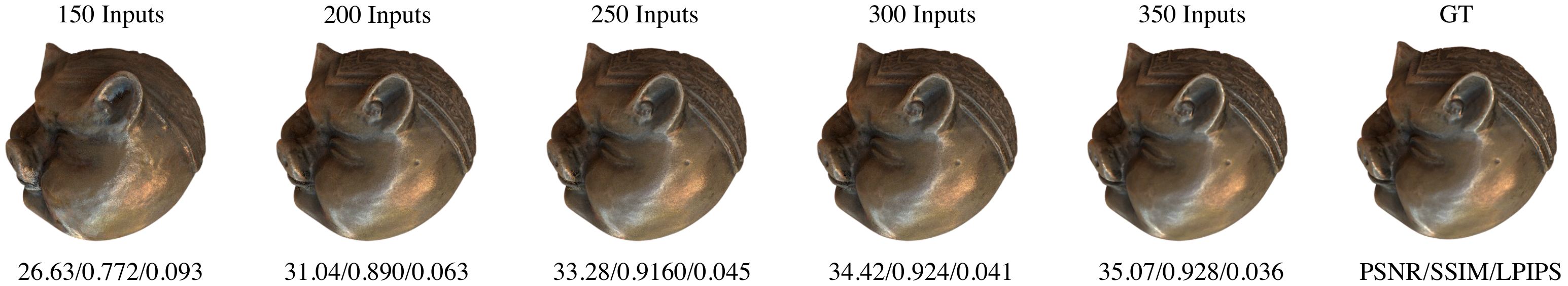}
	\caption{Ablation study on the number of input images. We show a novel view synthesis result of the reconstruction for each object, given a different number of input images, whose viewpoints are evenly distributed inside the upper hemisphere. \bf{We recommend that readers zoom in to see the details of the picture, especially the folds at the top and the specular highlights on the face.}}
	\label{4}
\end{figure*}

\paragraph{Real Data.}

We evaluate our method on multiple real-world images from the DTU datasets~\cite{aanaes2016large}, where the objects are glossy and the illumination is static across different views. We use two objects from the dataset, \emph{shiny scan114 buddha} and \emph{scan110 ghost}, and discard photos with strong shadows. 
Our inverse rendering results are shown in Figure \ref{3}. We can see that our pipeline generates photo-realistic novel views, and results of relighting and material editing.

\paragraph{Comparison with State-of-the-art Methods.}

As far as we are aware, there is no prior work tackling exactly the same problem as our work: reconstructing triangle meshes and PBR materials from multi-view real-world object images with uncontrolled lighting conditions. The methods closest to our method are~\cite{luan2021unified,li2021shape}, but no source code or data is released. It is worth noting that our method addresses the shortcomings of~\cite{li2021shape} in dealing with geometry, while overcoming the limitation of~\cite{luan2021unified} which requires the camera and point lighting to be in the same position and the images to be taken in dark scenes. 

We compare our method to the most related neural rendering approaches, including PhySG~\cite{zhang2021physg}, NeRF~\cite{mildenhall2020nerf} and IDR~\cite{yariv2020multiview}, in terms of novel view synthesis. These approaches are different from our method in the model of light transport: NeRF uses the occupancy function to describe geometry and appearance maps and gets the pixel color according to location and viewing direction in a ray-marching way, while PhySG and IDR use SDF to represent geometry and material. In addition, unlike the aforementioned two methods that focus solely on novel view synthesis, PhySG also performs inverse rendering tasks, using Spherical Gaussian functions to describe materials and illumination. 
Qualitative and quantitative comparisons between different methods on our synthetic data are depicted in Figure~\ref{comparative} and Table~\ref{comparative-table}. The unsatisfactory performance of PhySG may be due to the presence of strong specular highlights in our synthetic data, which are difficult to describe using the Spherical Gaussian functions employed by PhySG. NeRF performs relatively poorly in view synthesis because its volumetric representation does not concentrate colors around surfaces as well as surface-based approaches. IDR does a better job in view synthesis because of its view-dependence model. However, it still struggles to synthesize specular highlights 
due to the non-physical model of appearance. In contrast, our method models highlight well, benefiting from our PBR materials.
We also compare the running time, where our training time is the sum of initialization time and optimization time. Thanks to the combination of efficiency from the approximate method and the accuracy from the physically-based method, our hybrid approach runs significantly faster than the two baselines both in training and testing. In addition, our optimized geometry and material can be deployed directly on mainstream graphics engines.

\begin{figure}[!]
	\centering
	\includegraphics[width=\columnwidth]{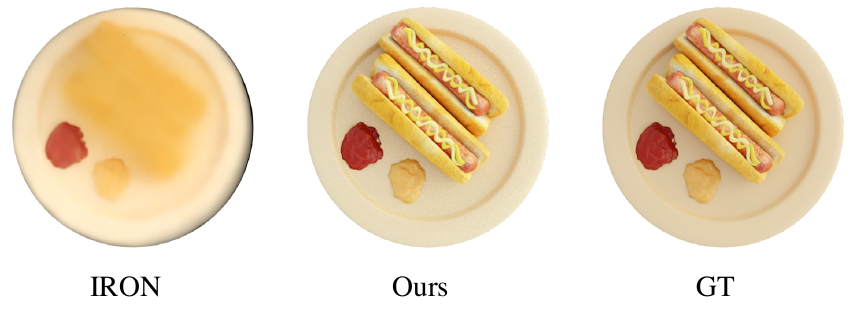}
        \caption{Comparison between our method and IRON~\protect\cite{zhang2022iron} on the hotdog data.}
	\label{IRON_hotdog}
\end{figure}

\begin{figure}
	\centering
	\includegraphics[width=0.99\columnwidth]{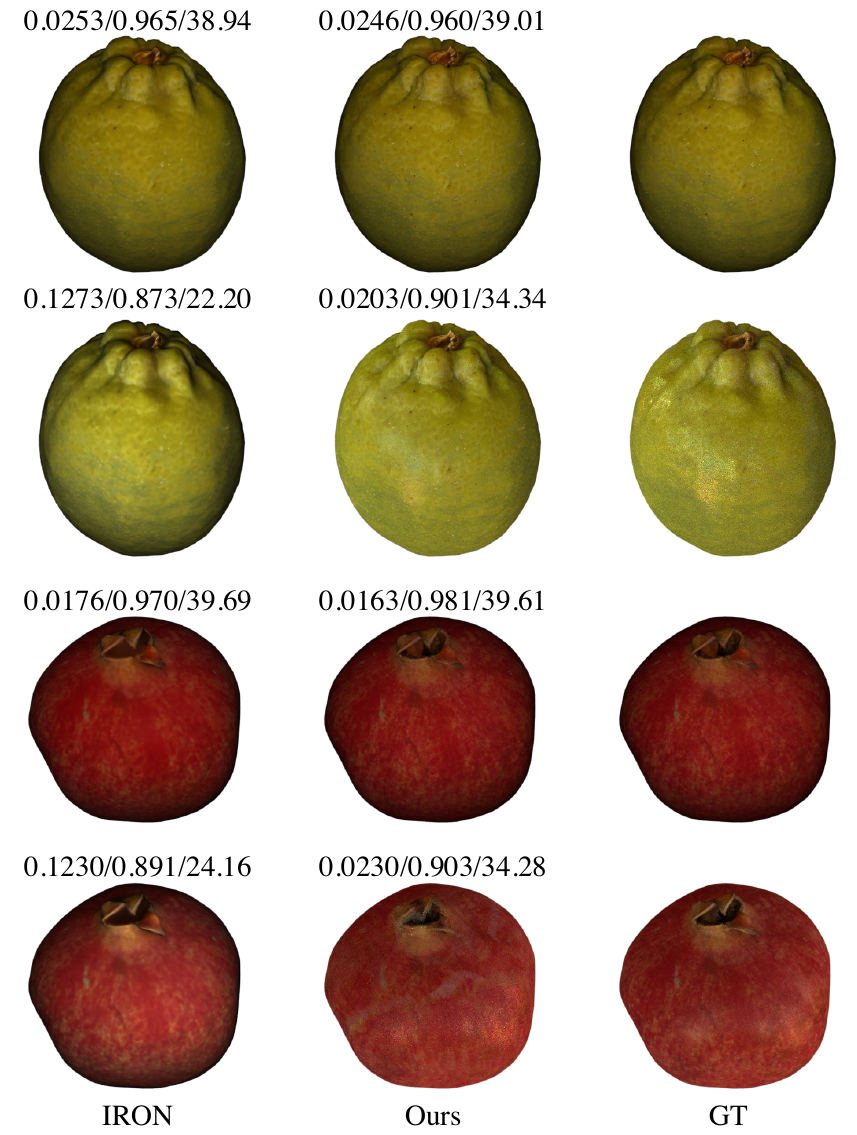}
        \caption{Comparison between our method and IRON~\protect\cite{zhang2022iron}. The first and third rows are point lighting, and the second and fourth rows are environment lighting. The numbers above each result show the  LPIPS, SSIM and PSNR in 512$\times$512 pictures, respectively.}
	\label{IRON}
\end{figure}

We also compare our method with the latest related work IRON~\cite{zhang2022iron} which adopts neural representations and also leverages a hybrid optimization scheme. Note that IRON assumes a point light source, while our method allows for uncontrolled lighting conditions. As a result, IRON performs poorly in complex lighting situations. Figure~\ref{IRON} compares the two methods in different lighting conditions.
Our method can achieve similarly good results as IRON on point-lighting data, while being superior with more complex lighting. In particular, IRON cannot handle the highlights on the surface of objects under environment lighting because their light source cameras are in the same position, while our method can handle it well. In addition, for some flat data like the hotdog in Figure~\ref{IRON_hotdog}, IRON struggles to obtain appropriate geometry, while our method can produce an accurate reconstruction.

\paragraph{Ablation Study.}

Figure~\ref{4} shows our ablation study on the number of input images, which depicts how the number of input images affects our reconstruction quality. With too few input images, the optimization may become highly under-constrained, making it difficult to produce accurate synthesis results. In our experiments, $150$ images are sufficient to produce a quite good result. With $250$ or more input images, our results will closely match the ground truth.
More ablation studies can be found in the supplementary materials.

\section{Conclusion}

We introduce a novel efficient hybrid differentiable rendering method to reconstruct triangular object mesh and PBR material from multi-view real-world images with uncontrolled lighting conditions. Unlike prior works that require massive resource consumption or approximated rendering process, we utilize an approximate method to optimize the geometry and a physically-based method to estimate the reflectance so that we could benefit from both the efficiency of the former and the high quality of the latter. In general, our method can handle a wide range of real-world scenes, providing an attractive and efficient solution and enabling photo-realistic novel view synthesis and relighting applications.

\paragraph{Limitations and future work.} Our method can have difficulties with very thin geometry, which is a common problem of mesh-based methods. In addition, our method optimizes geometry and material separately. A potential future work is a unified pipeline to optimize geometry and material simultaneously, which should further improve the result quality.

\section*{Acknowledgments}
This work was supported by the NSFC under Grant 62072271.

\bibliographystyle{named}
\bibliography{ijcai23.bib}

\end{document}